\documentclass[a4paper, 10pt, conference]{ieeeconf}      % Use this line for a4
\usepackage{algorithm,algorithmic, amsmath, amssymb, url}
\usepackage{multirow, booktabs, graphicx, subcaption}

\IEEEoverridecommandlockouts                              % This command is only
\usepackage{amsmath} % assumes amsmath package installed

\title{\LARGE \bf
Fair and Interpretable Deepfake Detection in Videos
}

\author{\parbox{16cm}{\centering
    {\large Akihito Yoshii$^{1*}$, Ryosuke Sonoda$^{1*}$, Ramya Srinivasan$^2$}\\
    {\normalsize
    $^1$ Fujitsu Limited, Japan\\
    $^2$ Fujitsu Research of America, Inc., USA}}
    \thanks{$^*$Equal contribution.}% <-this % stops a space
}

\begin{document}

\maketitle

%%%%%%%%%%%%%%%%%%%%%%%%%%%%%%%%%%%%%%%%%%%%%%%%%%%%%%%%%%%%%%%%%%%%%%%%%%%%%%%%
\begin{abstract}

Existing deepfake detection methods often exhibit 
bias, lack transparency, and fail to capture temporal information, leading to biased decisions and unreliable results across different demographic groups.
In this paper, we propose a fairness-aware deepfake detection framework that integrates temporal feature learning and demographic-aware data augmentation to enhance fairness and interpretability.
Our method leverages sequence-based clustering for temporal modeling of deepfake videos and concept extraction to improve detection reliability
while also facilitating interpretable decisions for non-expert users.
Additionally, we introduce a demography
-aware data augmentation method that balances underrepresented groups and applies frequency-domain transformations to preserve deepfake artifacts, thereby mitigating bias and improving generalization.
Extensive experiments on FaceForensics++, DFD, Celeb-DF, and DFDC datasets using state-of-the-art (SoTA) architectures (Xception, ResNet) demonstrate the efficacy of the proposed method in obtaining the best tradeoff between fairness and accuracy when compared to SoTA.

\end{abstract}

%%%%%%%%%%%%%%%%%%%%%%%%%%%%%%%%%%%%%%%%%%%%%%%%%%%%%%%%%%%%%%%%%%%%%%%%%%%%%%%%
\section{INTRODUCTION}

The rise of deepfakes has posed a major threat to the safety and privacy of individuals, institutions, societies, and nations \cite{mustak,farid}. Scholars posit that with the rapid proliferation of deepfakes, we are heading towards an ``infopocalypse" where we cannot tell what is real from what is not \cite{don}. To add to this threat is the fact that the very technologies that enable innovation can be manipulated for creation of deepfakes, resulting in malicious content that undermine privacy and promote disinformation \cite{zhiyuan}.

In response to these growing concerns, researchers and practitioners have developed a suite of deepfake detection methods that also generalize to out of distribution datasets, taking into account the multiple manipulations that deepfakes may undergo \cite{akash,yan,kartik}. In parallel, there have also been efforts to develop standardized, unified, and comprehensive benchmarks that enable fair comparison of various deepfake detection methods \cite{zhiyuan, chuqiao}. Despite these efforts, challenges remain. 

Preserving fairness in deepfake detection across demographic groups is one prominent challenge \cite{akshay}. Although recent methods such as \cite{li} have investigated this problem by proposing disentanglement learning to extract demographic and domain-agnostic forgery features to encourage fair learning, the method does not take into account spatio-temporal changes which can affect both accuracy and fairness of deepfake detectors.
Another challenge concerns effectively learning dynamic changes to uncover spatio-temporal manipulations. 
Although existing works such as \cite{Han_2025_CVPR} consider spacial manipulation cues and temporal inconsistency , such features are still prone to biases owing to their latent correlations with sensitive attributes such as race or gender.
A third challenge is to enable transparent deep fake detection and mitigation, whereby lay-users can understand the rationale behind fake identification. This requirement has become more important than before given the new regulations around AI such as the EU AI act among others \cite{eu}. Although explainable AI methods have been developed for applications such as facial affect detection \cite{xinyu, guanyu}, their feasibility in deepfake detection remains limited. 

\begin{figure}[ht]
    \centering
    \includegraphics[width=0.5\textwidth]{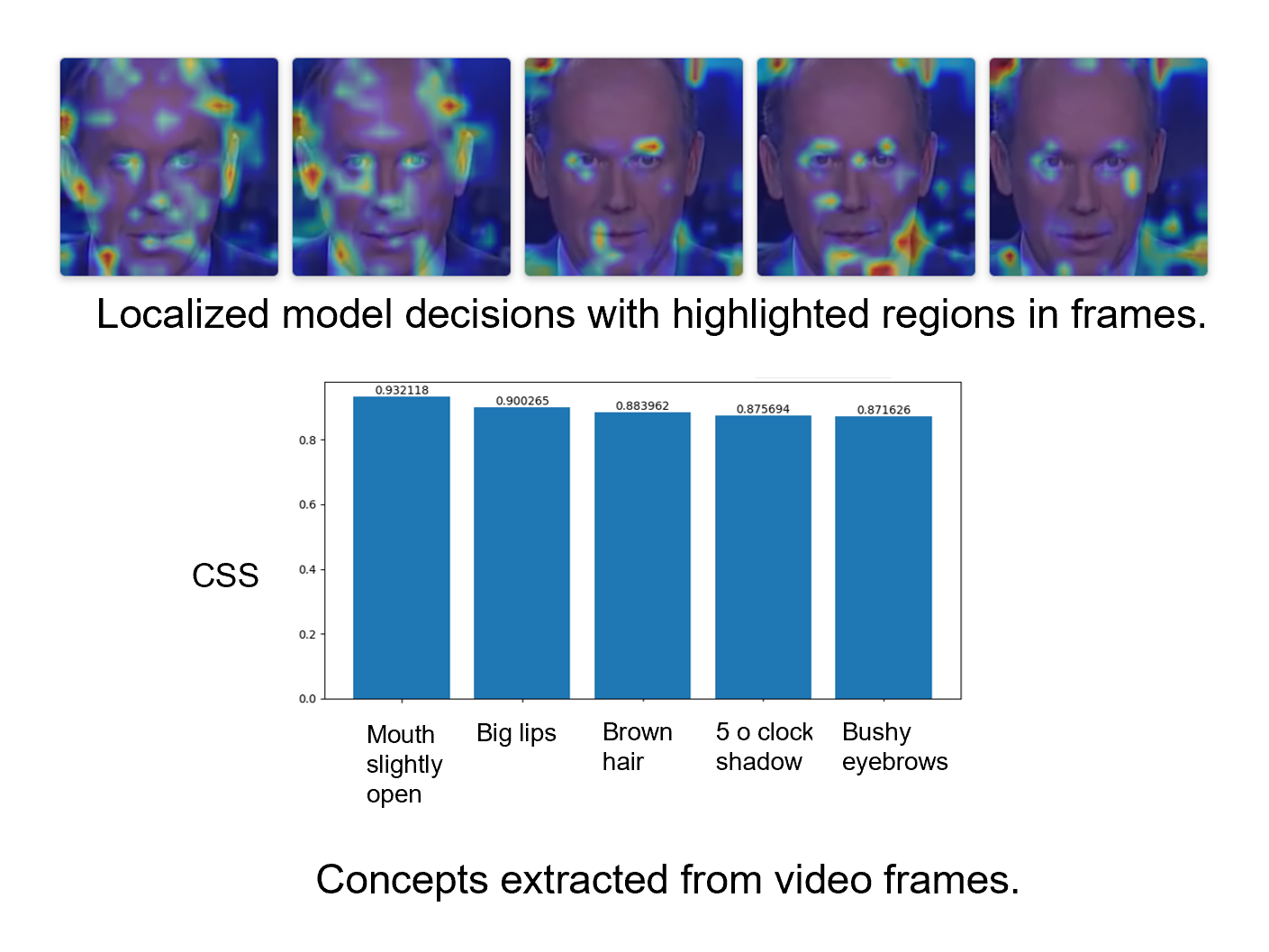}
    \caption{\textbf{Illustration of system outputs for an example input video.} Five frames with heat maps highlighting the likelihood of fake regions and Concept Sensitivity Score (CSS) providing human-interpretable explanations of model's decision. }
    \label{fig:explanation}
\end{figure}

\begin{figure*}[ht]
    \centering
    \begin{subfigure}{\textwidth}
        \centering
        \includegraphics[width=\textwidth]{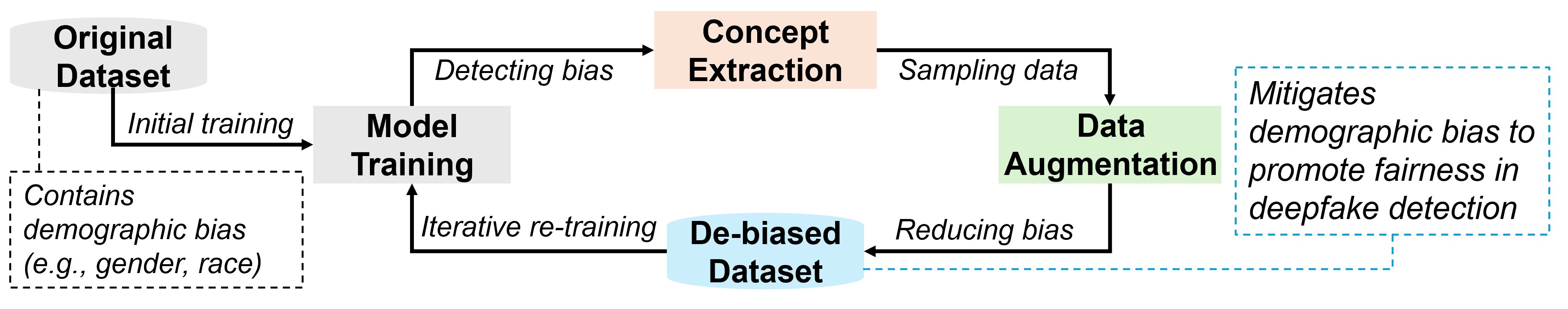}
        \caption{Overview of our framework. Our core contributions lie in the concept extraction and data augmentation modules, which can enhance model performance.}
        \label{fig:overview}
    \end{subfigure}
    %\hspace{0.1\textwidth}
    \begin{subfigure}{0.45\textwidth}
        %\centering
        \includegraphics[width=\textwidth]{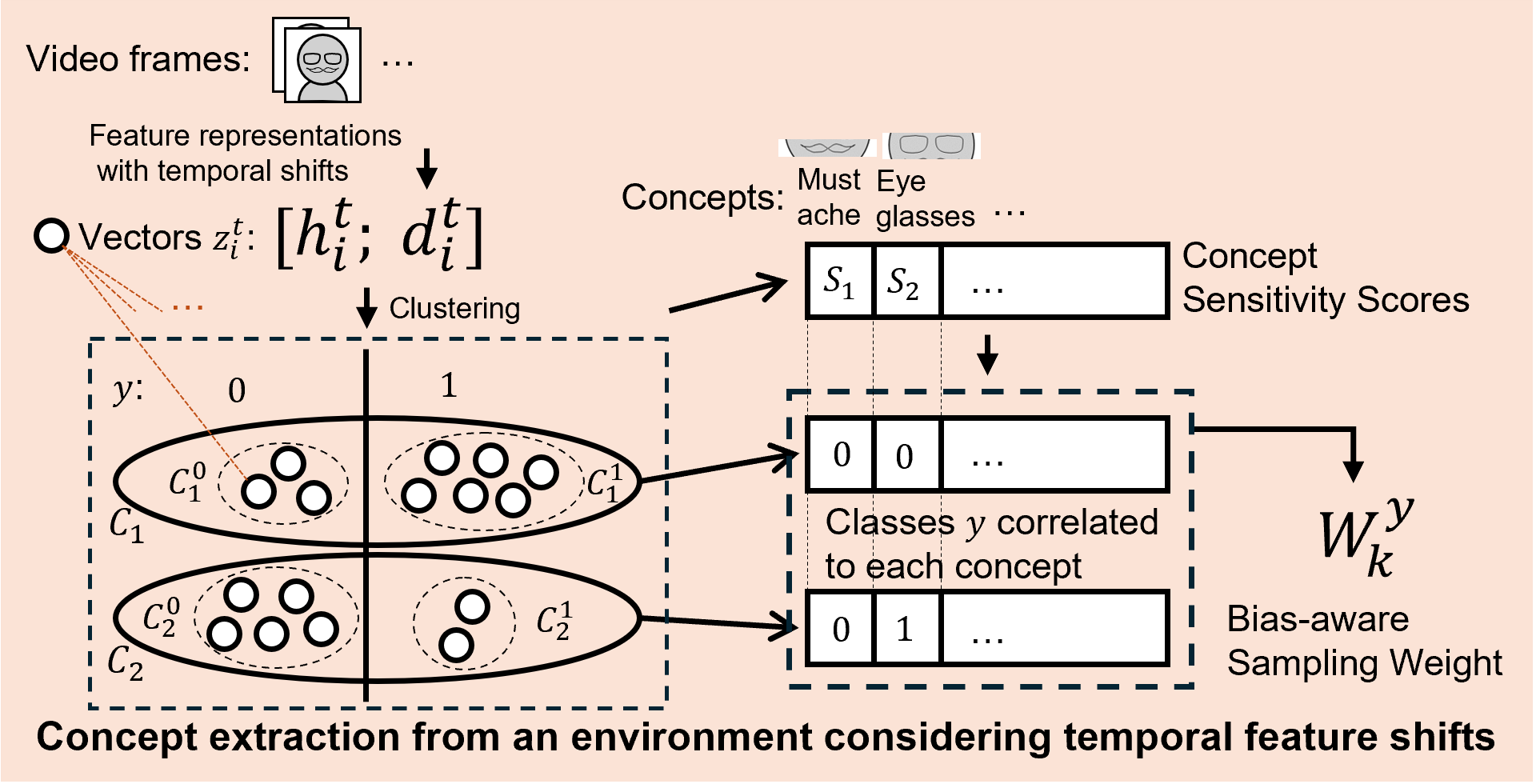}
        \caption{Illustration of the proposed concept extraction method.}
        \label{fig:system_concept}
    \end{subfigure}
    \hfill
    \begin{subfigure}{0.45\textwidth}
        %\centering
        \includegraphics[width=\textwidth]{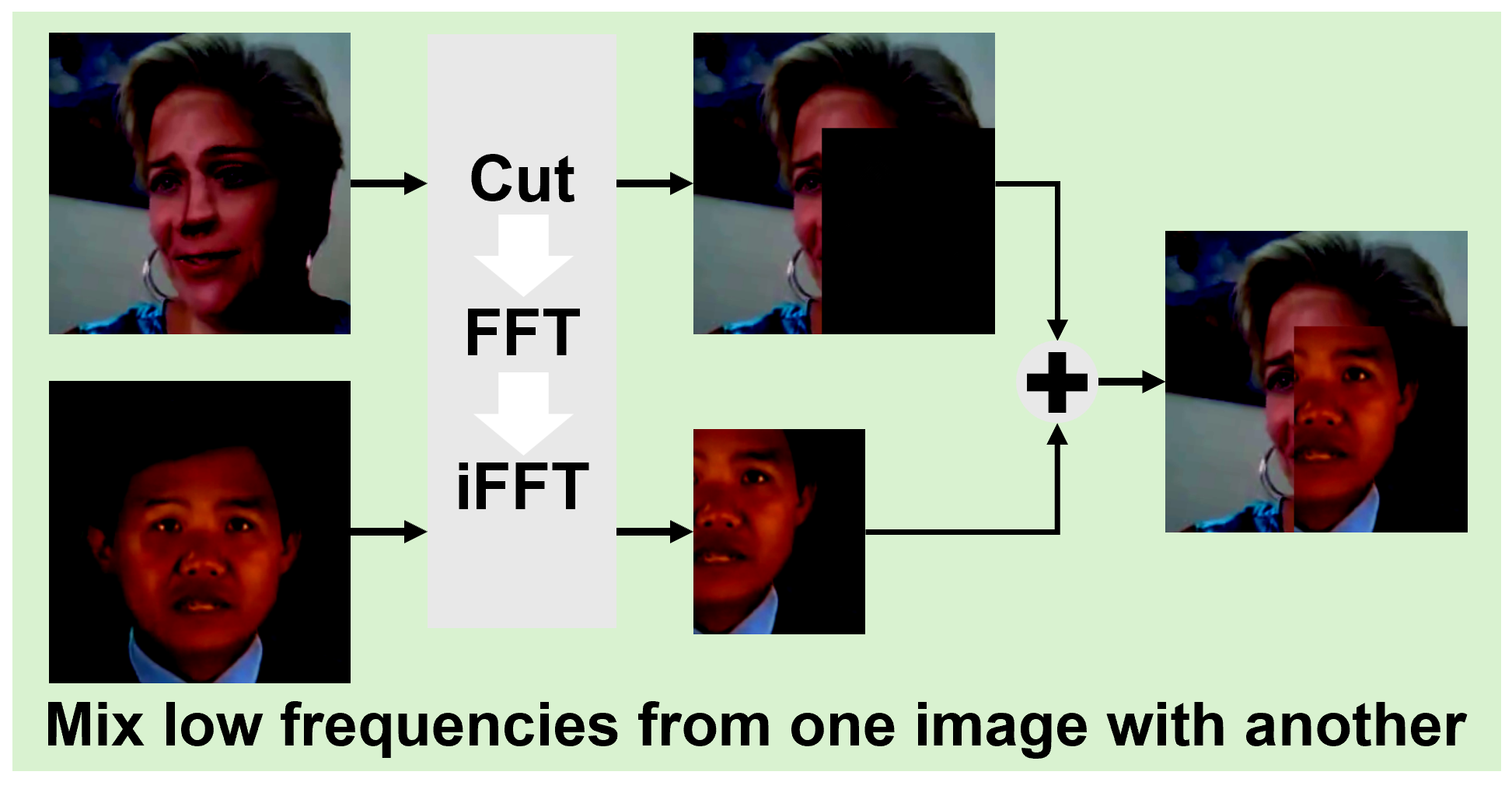}
        \caption{Illustration of the proposed data augmentation method.}
        \label{fig:system_augment}
    \end{subfigure}
    \caption{\textbf{System diagram.} Our framework first extracts proxy attributes for demographic attributes from the training data. Next, it applies frequency-aware data mixing to mitigate biases associated with these attributes. Finally, the model is re-trained on the de-biased dataset.}
    \label{fig:system_diagram}
\end{figure*}

{\bf Contributions:}
Towards addressing the aforementioned challenges, in this work, we develop a novel deep-fake detection method that can uncover subtle manipulations in videos while mitigating biases. The proposed framework leverages spatio-temporal cues to effectively detect minute manipulations across video frames. The proposed method offers fine-grained analysis by highlighting the fake regions in each frame in terms of human-interpretable concepts (e.g., facial mole, spectacle shape, etc.), thereby providing a user-friendly explanation and visualization  \cite{Selvaraju_2017_ICCV}(Fig.~\ref{fig:explanation}). Furthermore, the proposed framework also includes a novel frequency-aware data augmentation method that mitigates bias in deepfake detection across sensitive attributes such as gender and race. 
The proposed method takes into account high-frequency components of video frames where deepfake-specific artifacts are most prominent, ensuring no negative impact on model performance while promoting fairness in deepfake detection.
Extensive experiments on state-of-the-art face datasets demonstrate the effectiveness of the proposed methods.  Fig.~\ref{fig:system_diagram} provides an overview of the overall system.  

%%%%%%%%%%%%%%%%%%%%%%%%%%%%%%%%%%%%%%%%%%%%%%%%%%%%%%%%%%%%%%%%%%%%%%%%%%%%%%%%
\section{Related work}

In this section, we review recent works related to deepfake detection in videos. We also situate our work in the context of recent methods related to bias mitigation and transparency in deepfake detection. 
\subsection{Deepfake Detection in Videos}
A significant number of techniques to detect deepfakes in videos are based on deep learning methods. In \cite{rossler}, a deep convolutional neural network, known as XceptionNet, has demonstrated high accuracy in detecting deepfake videos. It was submitted to the DeepFake Detection Challenge (DFDC), receiving a score of 0.9965 for its AUC-ROC. In \cite{afchar}, the authors proposed a deep learning architecture called Mesonet to identify manipulated facial expressions. EfficientNet and ResNet based architectures have also proven to be effective in deepfake detection \cite{tan,he,agarwal}. More recently, transformer based models are also being employed for deepfake detection \cite{zhao} \cite{Cui_2025_CVPR}. On the other hand, \cite{Yan_2025_CVPR} leverages spatiotemporal features introducing an adapter applicable to existing models. As adopted by most state of the art techniques, we compare our method using ResNet and XceptionNet 
based architectures across multiple datasets to demonstrate the efficacy of the proposed methods. Further, unlike most of the existing works, the proposed method leverages both spatial and temporal information in not only detecting deepfakes but also in terms of enhancing stakeholders' understanding of the results.
\subsection{Bias Mitigation in Deepfake Detection}
Deepfake detection methods have shown varied performance across different genders and races, markedly showing higher false positive rates on certain minority groups \cite{ying}. The results from \cite{ying} showed that deepfake detection methods trained on such imbalanced/biased datasets result in incorrect detection results leading to generalizability, fairness, and security issues. In order to make detection results statistically independent of demographic factors and thereby improve fairness, the authors in \cite{ju} propose novel loss functions that handle both the setting where demographic information is available as well as the case where this information is absent.  Other methods include learning demographic-agnostic features \cite{li}, but their utility across datasets needs investigation. 
Beyond bias mitigation, data augmentation is widely used to enhance model generalization and robustness. While traditional methods such as MixUp~\cite{zhang} and CutMix~\cite{yun} improve generalization by blending training samples, these methods overlook frequency-specific deepfake artifacts, which can be crucial for detection.
Recent work explores frequency-aware augmentation, which modifies representations in the frequency domain rather than relying on spatial transformations~\cite{doloriel}. 
Additionally, synthetic datasets with balanced demographic representation have been proposed to improve fairness in deepfake detection~\cite{uzoamaka}.
Our framework introduces a frequency-aware, demographically balanced augmentation strategy that operates in the low-frequency domain, enhancing fairness while preserving deepfake-specific artifacts.

\subsection{Transparency in Deepfake Detection}
Recent studies have shown that explainable AI methods can enhance deepfake detection \cite{xhu,nazneen,rozsa}. In \cite{loc}, the authors propose a novel human-centered approach for detecting forgery in face images, using dynamic prototypes as a form of visual explanations. In \cite{gazi}, the authors utilize CNN (Convolutional Neural Network) and CapsuleNet with LSTM to differentiate between deepfake-generated frames and originals to aid users in identifying fake videos.  In \cite{shichao}, the authors interpret how deepfake detection models learn artifact features of images when just supervised by binary labels and demonstrate that deepfake detection models indicate real/fake images based on visual concepts that are neither source-relevant nor target-relevant, but rather artifact relevant. Motivated by these findings, in this work, we propose a complementary approach whereby we extract concepts that contain implicit demographic information and demonstrate the effectiveness of the proposed approach in mitigating biases across state-of-the-art deepfake detection datasets.

\addtolength{\textheight}{-3cm}   % This command serves to balance the column lengths
\section{Method}
Let $f(\cdot)$ be a deepfake detector trained on the dataset $\mathcal{T}_{\mathrm{train}} = \{(x_i^t, y_i^t)\}_{i=1}^{N}$, where each video $i$ consists of a sequence of frames $\{x_i^t\}_{t=1}^{T_i}$, and $y_i^t \in \{0,1\}$ denotes whether frame $t$ is real ($y_i^t=0$) or fake ($y_i^t=1$).
The performance of $f$ is evaluated on a test set $\mathcal{T}_{\mathrm{test}} = \{(x_j^t, y_j^t, a_j^t)\}_{j=1}^{N'}$, where $a_j^t$ represents a demographic attribute (e.g., race or gender) unavailable in $\mathcal{T}_{\mathrm{train}}$.
For simplicity, we denote a sampled frame as $x_i$ unless the time index $t$ is not explicitly required.
Our goal is to train a model $f$ on $\mathcal{T}_{\mathrm{train}}$ that enhances both accuracy and fairness when evaluated on $\mathcal{T}_{\mathrm{test}}$.

\noindent\textbf{Motivation.} Our work is motivated by the following observations--- i) Deepfake detection models often exhibit bias due to spurious correlations in low-frequency components or imbalances in demographic groups within the training set~\cite{nadimpalli,trinh}, leading to disparities in performance across demographic groups, ii) deepfake-specific artifacts often hide in the high-frequency domain of images~\cite{doloriel}, iii) labeling of demographic attributes in images can be expensive and is often unavailable ~\cite{ying}.

\subsection{Proposed Concept Extraction Method}\label{subsec:concept_extraction}
In the absence of labeled demographic attributes, we propose a 
concept-based approach to identify potential biases in deepfake detection models.
Rather than directly grouping feature representations using unsupervised techniques which may fail to capture demographic attributes, we extract high-level concepts that implicitly encode demographic information, such as skin tone, hairstyle, or accessories~\cite{5206594,pmlr-v119-koh20a}.
These inferred concepts allow for a more systematic analysis of demographic disparities 
in training dataset, providing insights into potential sources of bias. 

\subsubsection{Concept Bank Construction}
The concept bank is a structured repository of $L$ human-interpretable concepts, each represented by a set of images~\cite{wu}:
\begin{equation}
    \mathcal{C} = \{c_l\}_{l=1}^L,
\end{equation}
where each $c_l$ represents a concept. 
The concept bank serves as an external knowledge source for identifying spurious correlations in model predictions by explicitly linking feature representations to human-interpretable concepts.
To quantitatively represent concepts in the model $f$'s feature space, we define a high-dimensional concept representation vector $\mathbf{v}_l$ for each $c_l$. 
This vector is obtained by training a linear classifier (e.g., a Support Vector Machine) to distinguish images containing the concept from those that do not~\cite{kim2018interpretability}. 
The resulting classifier provides a separating hyperplane in the feature space, where the normal vector $\mathbf{v}_l$ represents the most discriminative direction for detecting the presence of concept $c_l$. 
By projecting model representations onto these concept vectors $\mathbf{v}_l$, we can analyze how specific concepts influence the model’s predictions.

\subsubsection{Identification of Concept-based Bias}
To identify biased concepts from the candidates in the concept bank, we introduce a clustering-based approach that leverages the model’s learned feature representations.
Our approach adopts the clustering procedure based on \cite{wu}. Unlike \cite{wu}, which assumes static input images, our method considers temporal differences between frames in a video.
\paragraph{Clustering with Temporal Information.}\label{para:clustering}
Deepfake videos exhibit temporal inconsistencies, which can provide additional cues for bias analysis.  
To account for this, we incorporate temporal differences 
into the clustering process.
Given a model $f$ trained on $\mathcal{T}_{\mathrm{train}}$, we extract a feature representation $\mathbf{h}_i^t$ for each training sample (video frame) $x_i^t$.
Since raw feature embeddings can be high-dimensional and computationally expensive for clustering, we apply dimensionality reduction techniques such as PCA or UMAP~\cite{mcinnes2020umapuniformmanifoldapproximation} to obtain a compact representation $\tilde{\mathbf{h}}_i^t$.  

To incorporate temporal variation, we define the temporal difference 
$d_i^t$ as:

\begin{align}
    d_i^t = 1 - \cos(\tilde{\mathbf{h}}_i^{t-1}, \tilde{\mathbf{h}}_i^t),
\end{align}

where $\cos(\tilde{\mathbf{h}}_i^{t-1}, \tilde{\mathbf{h}}_i^t)$ denotes the cosine similarity between the feature vectors of consecutive frames.  
$d_i^t$
measures the degree of feature shift between successive frames, with larger values indicating greater temporal inconsistency.  
For the first frame of a video ($t=0$), we define 
$d_i^0 = 0$
as there is no preceding frame.

Finally, we concatenate the temporal difference 
$d_i^t$ with the reduced feature representation $\tilde{\mathbf{h}}_i^t$ to construct the final clustering input:
\begin{equation}\label{eq:clustering_vector}
    \mathbf{z}_i^t = [\tilde{\mathbf{h}}_i^t; d_i^t].
\end{equation}
By incorporating $d_i^t$, our clustering approach accounts for both spatial feature similarity and temporal inconsistency, which enhances the identification of bias-inducing patterns in deepfake detection. Given these feature representation $\mathbf{z}_i^t$, we cluster the data within each class.  
For each class $ y \in \{0,1\}$, we obtain $K$ disjoint clusters, denoted as $ C_1^y, \dots, C_K^y$. 

\paragraph{Quantifying Bias Using Concept Sensitivity Score (CSS).}  
After clustering, we measure the extent to which each concept exhibits pseudo/spurious correlation with class labels.  
For this, we employ the Concept Sensitivity Score (CSS), adapted from~\cite{wu}, which quantifies how inconsistently a concept is distributed across different clusters.

To measure CSS, we first define the \textit{environment} $C_k$, which consists of merged clusters containing samples from both classes (i.e., real and fake):
\begin{equation}\label{eq:environment}
     C_k = C_{k_0}^0 \cup C_{k_1}^1, \quad k_0, k_1=1,\dots,K.
\end{equation}
Unlike individual clusters that are inherently class-dependent, environments $C_k$ group samples from both classes together, allowing us to analyze concept behavior under diverse conditions. 
By examining how the presence of a concept fluctuates across environments, we can identify inconsistencies in its association with class labels.
For example, if ``pale skin" exhibits highly variable distributions across environments, it may indicate unintended demographic bias. 
To enhance robustness, the formulation of environments can be randomized during training, as suggested in~\cite{wu}. 

Using the environment $C_k$, we define the CSS as:
\begin{equation}\label{eq:css}
    S_l = \text{Var} \left( \{(\mathbf{v}_l \cdot \mathbf{M}_k^T)_{y_l'} \mid k=1, \dots, K \} \right),
\end{equation}
where  
\begin{itemize}
    \item $\mathbf{M}_k = \nabla_\theta \left[\mathbb{E}_{(x,y) \sim C_k} \mathcal{L}(f(x),y) \right]$ is the gradient matrix of the model loss $\mathcal{L}(\cdot)$ w.r.t. parameters $\theta$ in environment $C_k$.
    \item \( y_l' = \arg\max_y \sum_k \mathbf{v}_l \cdot \mathbf{M}_k^T \) is the class most strongly associated with concept $c_l$.
    \item \( \text{Var}(\cdot) \) denotes variance.
\end{itemize}
Intuitively, a high CSS value indicates that a concept's association with class labels is inconsistent across different environments, suggesting spurious correlations.  
For example, if ``bald" frequently co-occurs with ``fake" in training data, but not in all clusters, its CSS would be high, indicating an unreliable correlation.
Since CSS is computed separately for each
concept, it provides fine-grained interpretability, helping in identifying class-specific biases.

\subsubsection{Bias-aware Sampling Strategy}
Once biases are quantified, we propose a bias-aware sampling strategy for data augmentation method to mitigate their impact. 
The idea is to re-balance the training data distribution by adjusting the sampling probabilities based on detected biases. 

A simple yet effective approach is to sample inversely proportional to the cluster size $|C_k^y|$, ensuring that minority clusters receive higher sampling probability.  
Formally, the base sampling weight is defined as:
\begin{equation}
    r(k,y) = \frac{1}{|C_k^y|}.
\end{equation}
The probability of selecting a sample from cluster $C_k^y$ is then given by:
\begin{equation}\label{eq:proportional_sampling}
    P_{\text{size}}(k,y) = \frac{r(k,y)}{\sum_{k',y'} r(k',y')}.
\end{equation}
This formulation ensures that samples from smaller clusters are drawn more frequently, reducing the imbalance in training data distribution.

Beyond cluster size, we consider the degree of spurious correlations within each cluster using the masked Concept Sensitivity Score (MCSS).
For each concept $l$, its MCSS within class $y$ is defined as:
\begin{equation}
     S_l^y = S_l \cdot H_l^y,
\end{equation}
where $H_l^y$ is a binary mask such that $H_l^y = 1$ if $y_l' = y_l$, and $H_l^y = 0$ otherwise.
To calculate probability of a concept's correlation with a class, we define MCSS probability:

\begin{equation}
    P_{\text{concept}}(l,y) = \frac{S_l^y}{\sum_{l\in L_{y}} S_l^y},
\end{equation}
where $L_y$ denotes the set of concept appearing in class $y$. Among concepts that are strongly correlated with a class $y$, a concept with higher MCSS probability is given more weight in the calculation of eq. \ref{eq:proposed_sampling}.

To quantify the overall MCSS probability for a given cluster $C_k^y$, we aggregate MCSS probability over all concepts present in the cluster as follows:
\begin{equation}
    S(k,y) = P(\bigcup_{l\in L_{k,y}}A_{\text{concept}}(l,y))
\end{equation}

where $L_{k,y}$ denotes the set of concepts appearing in cluster $C_k^y$ and $A_{\text{concept}}(l,y)$ is an event with  probability $P_{\text{concept}}(l,y)$. $S(k,y)$ corresponds to the probability of a sum event through $L_{k,y}$.
A higher $S(k,y)$ value indicates stronger spurious correlations, suggesting greater potential bias.

To jointly account for representational imbalance and concept-based bias, we propose a bias-aware sampling weight:
\begin{equation}\label{eq:proposed_sampling}
     W(k,y) = S(k,y) \cdot r(k,y).
\end{equation}
This weighting scheme ensures that sampling prioritizes clusters that are both underrepresented and exhibit higher degrees of bias, leading to a more balanced and de-biased training distribution. 
Thus, our bias-aware sampling strategy effectively counteracts both data imbalance and spurious correlations, promoting fairer and more robust model training.

\paragraph{Connection with Data Augmentation}  
Existing method to learn CSS ~\cite{wu} employs bias-free sampling strategy with common data augmentation methods such as MixUp ~\cite{zhang} and CutMix ~\cite{yun}.
However, these methods are not tailored for fair deepfake detection and do not necessarily help in mitigating biases in the model.
In the next section, we present a novel data augmentation method specifically designed to address biases in deepfake detection.

\begin{algorithm}[t]
\caption{Proposed framework}\label{algorithm}
\textbf{Input:} Training data $\mathcal{T}_{\mathrm{train}}$, a model $f$, batch size $N_b$, cluster size $K$, a concept bank $\mathcal{C}$ \\
\textbf{Output:} A fair deepfake detector
\begin{algorithmic}[1]
\STATE Train $f$ on $\mathcal{T}_{\mathrm{train}}$
\STATE Obtain $K$ clusters using vectors defined in \eqref{eq:clustering_vector}
\WHILE{not converge}
    \STATE Sample a mini-batch $\mathcal{B} = \{(x_i,y_i)\}_{i=1}^{N_b}$ from $\mathcal{T}_{\mathrm{train}}$
    \STATE Construct environments from $\mathcal{B}$ using \eqref{eq:environment}
    \STATE Calculate CSS with the environments using \eqref{eq:css}
    \STATE Sample pairs of $(x_i,x_j)$ using \eqref{eq:proposed_sampling}
    \STATE Conduct data augmentation on the pairs to obtain $\mathcal{B}' = \{(x_i',y_i)\}_{i=1}^{N_b}$ using \eqref{eq: cutmix}
    \STATE Update $f$ with $\mathcal{B}'$
\ENDWHILE
\end{algorithmic}
\end{algorithm}

\subsection{Proposed Data Augmentation Method}\label{subsec:data_augmentation}
To mitigate bias and preserve deepfake-specific artifacts, we introduce a frequency-aware augmentation method that selectively modifies low-frequency components while retaining high-frequency artifacts. 
As shown in Fig.~\ref{fig:system_augment}, our augmentation method generates de-biased training data by selectively mixing low frequency components of different video frames while ensuring demographic diversity.

Let $(x_i, x_j) \in \mathcal{T}_{\text{train}}$ be a pair of training images, where $x_j$ is sampled according to the probability $W(k, y_i)$ defined in \eqref{eq:proposed_sampling}.
To generate the augmented sample $x'$, a region of $x_i$ in the low-frequency domain is replaced with the corresponding region from $x_j$. The resulting transformation is defined as follows: 
\begin{equation}\label{eq: cutmix}
    x' = \mathbf{M}_{\text{cut}} \odot \mathcal{LF}(x_i)+ \mathcal{HF}(x_i) + (\mathbf{1} - \mathbf{M}_{\text{cut}}) \odot \mathcal{LF}(x_j) ,
\end{equation}
where $\mathbf{M}_{\text{cut}} \in \{0,1\}^{H \times W}$ is a binary mask that determines the region to be mixed, sampled uniformly over a square patch within the spatial domain. 
Here, $\mathbf{1}$ is an all-ones matrix of the same dimension, and the $\odot$ denotes element-wise multiplication.
The function $\mathcal{LF}(x)$ extracts the low-frequency component of an image, while the term $\mathcal{HF}(x_i) =x_i-\mathcal{LF}(x_i)$ reconstructs the high-frequency component, ensuring that the original high-frequency details of $x_i$ remain intact.
This formulation preserves critical high-frequency artifacts essential for deepfake detection while mitigating biases present in the low-frequency domain.

Here we define the frequency decomposition of an image $x$ using a low-pass filter $\mathcal{LF}(\cdot)$.
We first compute the 2D Fast Fourier Transform (FFT) of the image, denoted as $\mathcal{F}(x)$. 
FFT converts an image of spatial dimensions $H \times W$ , where $H$ is the height and $W$ is the width, respectively, 
into the frequency domain, where the image is represented in frequency components $u$ and $v$, corresponding to the vertical and horizontal frequency components, respectively. 
The low-frequency components are then separated using a frequency mask $\mathbf{M}_{\text{low}}$:
\begin{equation}
    \mathcal{F}_{\text{low}}(x) = \mathcal{F}(x) \odot \mathbf{M}_{\text{low}},
\end{equation}
where the low-frequency mask $\mathbf{M}_{\text{low}}$ is defined as:
\begin{equation}
    \mathbf{M}_{\text{low}}(u, v) =
    \begin{cases}
        1, & \text{if } 0 \leq u < \alpha H , 0 \leq v < \alpha W, \\
        0, & \text{otherwise},
    \end{cases}
\end{equation}
where $\alpha$ is a hyperparameter that controls the size of the low-frequency region: when $\alpha = 1$, the entire image is considered as part of the low-frequency region, and when $\alpha = 0$, no low-frequency components are retained. In our experiments, we set $\alpha = 3/4$.
Applying the inverse FFT, we obtain the spatial domain representations, where low frequencies are retained while high frequencies are attenuated:
\begin{equation}
    \mathcal{LF}(x) = \mathcal{F}^{-1}(\mathcal{F}_{\text{low}}(x))
\end{equation}
Our method ensures demographic balance by selectively blending data with different demographic attributes, while applying augmentation in the low-frequency domain for the same class. 
This preserves high-frequency deepfake artifacts and minimizes the negative impact on performance in terms of drop in detection accuracy and model fairness. 
As a result, the model achieves balanced performance across demographic groups and enhances generalization to unseen deepfake operations.

The overall training procedure is detailed in Algorithm~\ref{algorithm}.
%%%%%%%%%%%%%%%%%%%%%%%%%%%%%%%%%%%%%%%%%%%%%%%%%%%%%%%%%%%%%%%%%%%%%%%%%%%%%%%%
\section{Experiment}
We begin by describing the experimental settings. 
\subsection{Experimental Settings}
\begin{table}[t]
    \centering
    \begin{tabular}{c|ccc} \hline
       Dataset  & \# Train & \# Validation & \# Test \\ \hline
        FF++ & 76,139 & 25,386 & 25,401\\
        DFD  & - & - & 9,385  \\
        DFDC & - & - & 22,857\\ 
        Celeb-DF & - & - & 28,458 \\ \hline
    \end{tabular}
    \caption{Number of samples in each dataset. ``-" means not used.}
    \label{tab:dataset}
\end{table}
\noindent\textbf{Datasets.} To assess both accuracy and fairness, we conduct training on the widely used FaceForensics++ (FF++)~\cite{faceforensics} dataset and evaluate performance on FF++, Deepfake Detection (DFD)~\cite{dfd}, Deepfake Detection Challenge (DFDC)~\cite{dfdc}, and Celeb-DF~\cite{celebdf}. 
As demographic attributes are not inherently available in these datasets, we follow established pre-processing and demographic annotation methods~\cite{ying}. 
Our study considers eight intersectional demographic groups categorized by gender and race: Male-Asian, Male-White, Male-Black, Male-Others, Female-Asian, Female-White, Female-Black, and Female-Others.
For face detection and alignment, we use Dlib~\cite{davis}, resizing detected faces to $256 \times 256$ for training and evaluation.
Table~\ref{tab:dataset} summarizes the dataset statistics.

\noindent\textbf{Evaluation Metrics.} To quantify detection performance, we utilize the Area Under the Curve (AUC) metric, in alignment with prior deepfake detection study~\cite{zhiyuan}. For fairness assessment, we employ three complementary metrics: Equal False Positive Rate $F_{\text{FPR}}$, Equal True Positive Rate $F_{\text{TPR}}$, and Equalized Odds $F_{\text{EO}}$, consistent with existing studies~\cite{ju,li}. The mathematical definition of those three fairness metrics are 
\begin{align}
F_{\text{FPR}} &:= \max_{a \in \mathcal{A}} \left\{\frac{\sum_i \mathbb{I}_{[\hat{y}_i=1, a_i=a, y_i=0]}}{\sum_i \mathbb{I}_{[a_i=a, y_i=0]}} - \frac{\sum_i \mathbb{I}_{[\hat{y}_i=1, y_i=0]}}{\sum_i \mathbb{I}_{[y_i=0]}} \right\}, \nonumber\\
F_{\text{TPR}} &:= \max_{a \in \mathcal{A}} \left\{\frac{\sum_i \mathbb{I}_{[\hat{y}_i=1, a_i=a, y_i=1]}}{\sum_i \mathbb{I}_{[a_i=a, y_i=1]}} - \frac{\sum_i \mathbb{I}_{[\hat{y}_i=1, y_i=1]}}{\sum_i \mathbb{I}_{[y_i=1]}} \right\}, \nonumber\\
F_{\text{EO}} &:= \max_{y \in \mathcal{Y}, a \in \mathcal{A}} \left\{\frac{\sum_i \mathbb{I}_{[\hat{y}_i=1, a_i=a, y_i=y]}}{\sum_i \mathbb{I}_{[a_i=a, y_i=y]}} - \frac{\sum_i \mathbb{I}_{[\hat{y}_i=1, y_i=y]}}{\sum_i \mathbb{I}_{[y_i=y]}} \right\},
\end{align}
where $\hat{y}$ is a model prediction and $\mathbb{I}_{[x]}$ is an indicator function that equals $1$ if $x$ is true, and $0$ otherwise.

\noindent\textbf{Baseline Methods.} We benchmark our approach against SoTA fairness-aware deepfake detection techniques, including DISC~\cite{wu} and demographic-aware-deepfake-detection (DAW-FDD)~\cite{ju}, as well as a Vanilla baseline, defined as a standard model trained without any fairness methods. 

To provide a more fine-grained analysis, we further compare different variants of key components within our proposed method and DISC. We investigate the following:
\begin{itemize}
    \item \textbf{Clustering Strategy:} We compare our proposed clustering method (\textsf{PC}) with naive clustering based on Gaussian Mixture Model (\textsf{NC}). Unlike \textsf{NC}, which applies conventional clustering techniques to the model’s feature representations, \textsf{PC} incorporates temporal difference vectors (as described in Section~\ref{subsec:concept_extraction}) to enhance bias identification.
    \item \textbf{Concept Inference Technique:} We leverage Concept Bank (\textsf{CB}) \cite{wu} for inferring the concepts and compare this setup with scenarios when concepts are not inferred (VariantB and C in Table 3).
    \item \textbf{Pair Sampling Strategy:} We evaluate our proposed bias-aware sampling strategy (\textsf{BS}) against the proportional sampling strategy (\textsf{PS}).
    \textsf{PS} samples data from minority cluster as defined in \eqref{eq:proportional_sampling}
    whereas \textsf{BS} aims to mitigate spurious correlations by re-balancing the training distribution.
    \item \textbf{Data Augmentation Method:} We compare our proposed frequency-based data augmentation method (\textsf{PF}) with other data augmentation methods, namely, MixUp (\textsf{MU})~\cite{zhang}, CutMix (\textsf{CM})~\cite{yun}, and Frequency Masking (\textsf{FM})~\cite{doloriel}. \textsf{MU} and \textsf{CM} are widely used augmentation strategies that blend image pairs to improve model generalization. \textsf{FM} is a recent technique that applies frequency-domain masking to enhance deepfake detection performance. \textsf{PF} is our proposed augmentation method, designed to further improve fairness in deepfake detection.
\end{itemize}

Additionally, we compare performance with standard architectures used in deepfake detection methods—ResNet34\footnote{\url{https://pytorch.org/hub/pytorch_vision_resnet/}} 
and
Xception\footnote{\url{https://data.lip6.fr/cadene/pretrainedmodels/xception-b5690688.pth}}--each trained using cross-entropy loss.

\begin{table*}[t]
    \centering
    \resizebox{\textwidth}{!}{
    \begin{tabular}{lcccccc|ccccc}
        \toprule
        \multirow{2}{*}{Dataset} & \multirow{2}{*}{Method} & \multicolumn{5}{c|}{Xception} & \multicolumn{5}{c}{ResNet-34} \\
        \cmidrule(lr){3-7} \cmidrule(lr){8-12}
        & & $F_{\text{FPR}}$ & $F_{\text{EO}}$ & $F_{\text{TPR}}$ & F1 score & AUC & $F_{\text{FPR}}$ & $F_{\text{EO}}$ & $F_{\text{TPR}}$ & F1score & AUC \\ 
        \midrule
        \multirow{4}{*}{FF++} & Vanila & 0.44 & \textbf{0.19} & 0.07 & \textbf{0.95} & 0.94 & 0.66 & 0.37 & 0.08 & 0.94 & 0.93 \\
        & DAW-FDD & 0.60 & 0.30 & \textbf{0.03} & \textbf{0.95} & \textbf{0.95} & 0.80 & 0.39 & 0.15 & 0.93 & 0.90 \\
        & DISC & 0.36 & 0.19 & 0.06 & 0.94 & 0.93 & 0.52 & 0.27 & \textbf{0.04} & \textbf{0.94} & \textbf{0.94} \\
        & Ours & \textbf{0.35} & \textbf{0.18} & 0.06 & \textbf{0.95} & \textbf{0.95} & \textbf{0.47} & \textbf{0.25} & 0.07 & 0.93 & 0.92 \\
        \midrule
        \multirow{4}{*}{DFDC} & Vanila & 0.33 & 0.27 & 0.64 & 0.46 & 0.59 & 0.37 & 0.67 & 0.97 & 0.62 & 0.53 \\
        & DAW-FDD & 0.42 & 0.35 & 0.88 & \textbf{0.60} & 0.58 & \textbf{0.19} & 0.27 & 0.42 & 0.61 & \textbf{0.57} \\
        & DISC & \textbf{0.26} & 0.38 & 0.60 & 0.51 & 0.59 & 0.29 & 0.50 & 0.83 & 0.60 & \textbf{0.57} \\
        & Ours & 0.32 & \textbf{0.27} & \textbf{0.47} & 0.55 & \textbf{0.60} & 0.22 & \textbf{0.21} & \textbf{0.28} & \textbf{0.65} & \textbf{0.57} \\ 
        \midrule
        \multirow{4}{*}{Celeb-DF} & Vanila & 0.47 & 0.29 & \textbf{0.72} & 0.63 & 0.62 & 0.34 & 0.45 & \textbf{0.91} & 0.65 & 0.60 \\ 
        & DAW-FDD & \textbf{0.25} & 0.42 & 0.94 & 0.72 & 0.58 & 0.26 & 0.50 & 0.94 & 0.73 & 0.61 \\ 
        & DISC & 0.37 & 0.37 & 0.91 & \textbf{0.74} & 0.63 & \textbf{0.21} & \textbf{0.45} & 0.96 & \textbf{0.76} & \textbf{0.64} \\ 
        & Ours & 0.42 & \textbf{0.35} & \textbf{0.90} & 0.73 & \textbf{0.65} & 0.37 & 0.48 & \textbf{0.91} & 0.73 & \textbf{0.63} \\ 
        \midrule
        \multirow{4}{*}{DFD} & Vanila & 0.53 & 0.28 & 0.13 & 0.90 & 0.79 & 0.24 & 0.15 & 0.10 & 0.88 & 0.78 \\
        & DAW-FDD & \textbf{0.41} & \textbf{0.21} & \textbf{0.06} & \textbf{0.94} & \textbf{0.82} & 0.51 & 0.37 & 0.28 & 0.81 & 0.64 \\
        & DISC & 0.51 & 0.27 & 0.11 & 0.91 & \textbf{0.82} & \textbf{0.33} & \textbf{0.17} & \textbf{0.07} & 0.91 & 0.77 \\ 
        & Ours & 0.49 & 0.26 & 0.09 & 0.92 & \textbf{0.82} & 0.39 & 0.20 & 0.08 & \textbf{0.92} & \textbf{0.79} \\
        \bottomrule
    \end{tabular}}
    \caption{Comparison with different methods in terms of accuracy and fairness on FF++, DFDC, Celeb-DF, and DFD. Higher values are preferred in accuracy and lower values for fairness. \textbf{Bold} indicates the best performance. }
    \label{tab:results}
\end{table*}

\noindent\textbf{Implementation Details.} To ensure a fair comparison across all experiments, we maintain a consistent set of hyperparameter values of batch size, training epochs, and optimizer throughout the training procedure. Specifically, all models are trained using a batch size of $64$ for a total of $10$ epochs. 
Optimization is performed using the Adam optimizer, with the learning rate fixed at $ \beta = 2 \times 10^{-4} $.
For DISC and our method, concept bank was constructed from generated concept sample images using Stable Diffusion model ~\cite{Rombach_2022_CVPR}. Prompts were constructed concatenating fixed keyword ``face''  with  forty pre-defined label names from CelebA\cite{liu2015faceattributes} metadata.
Two hundred concept images were generated for each concept label. 
The number of cluster size for each class was set as 4 in all experiments.

\subsection{Results}

\subsubsection{Performance comparison}

Table~\ref{tab:results} compares our method with SoTA methods, demonstrating its improved fairness generalization and detection performance. Using Xception architecture, the proposed method achieves the best AUC on all four deepfake detection datasets. Similar performance can also be observed with regards to ResNet architecture, thus ensuring no performance degradation even with less computational resources. 
To assess the consistency of performance gains, we conducted a Spearman rank correlation test comparing each baseline to our method. 
All comparisons yielded strong correlations ($\rho > 0.84$) and statistically significant p-values ($p < 10^{-11}$), confirming that the observed improvements are consistent and statistically significant.

Thus, the proposed method consistently outperforms baselines, achieving the best balance between fairness and accuracy. 

\subsubsection{Ablation Studies}
We evaluate the effectiveness of the two main modules of our system (the concept extraction module and data augmentation module) through ablation studies. 

\noindent\textbf{Effect of concept extraction.} We conduct ablation studies with regards the clustering approach employed and the data sampling strategy. 
 Table~\ref{tab:ablation_1}  shows performance comparison between these variants. 
 To examine the effectiveness of clustering, VariantB and VariantC are studied. The results shows that proposed clustering (\textsf{PC}) method improves AUC by 3\% except for FF++ and improves $F_{\text{EO}}$ by 4\% for FF++ and DFD, suggesting the effectiveness of the PC.
In conjunction with concept bank and bias aware sampling (\textsf{BS}), the proposed clustering method yields better $F_{\text{EO}}$ by 50\% on Celeb-DF and AUC is improved by 1\% with DFD, thus confirming its effectiveness.
To assess the robustness of these gains, we conducted Spearman rank correlation tests comparing each variant to our method. All comparisons yielded statistically significant results ($p < 0.01$), confirming that the observed improvements are consistent.

\begin{table*}[t]
    \centering
    \small
    \setlength{\tabcolsep}{4pt}
    \renewcommand{\arraystretch}{1.2}
    \begin{tabular}{lccccccccccc}
        \toprule
        \multicolumn{4}{c}{\multirow{2}{*}{Component}} & \multicolumn{8}{c}{Dataset} \\
        \cmidrule(l){5-12}
         &  &  &  & \multicolumn{2}{c}{FF++} & \multicolumn{2}{c}{DFDC} & \multicolumn{2}{c}{Celeb-DF} & \multicolumn{2}{c}{DFD} \\
        \cmidrule(lr){1-4} \cmidrule(lr){5-6} \cmidrule(lr){7-8} \cmidrule(lr){9-10} \cmidrule(l){11-12}
        Name & Cl & Cg & Ps & $F_{\text{EO}}$ & AUC & $F_{\text{EO}}$ & AUC & $F_{\text{EO}}$ & AUC & $F_{\text{EO}}$ & AUC\\
        \midrule
        VariantA & \textsf{NC} & \textsf{CB} & \textsf{BS}  & \textbf{0.17} & 0.95 & 0.30 & 0.59 & \textbf{0.35} & 0.63 & 0.27 & 0.82 \\
        VariantB & \textsf{NC} & \textbf{-} & \textsf{PS}  & 0.20 & 0.95 & 0.33 & 0.61 & 0.46 & 0.61 & \textbf{0.16} & 0.83 \\
        VariantC & \textsf{PC} & \textbf{-} & \textsf{PS}  & 0.25 & \textbf{0.96} & 0.37 & 0.61 & 0.41 & 0.64 & 0.20 & \textbf{0.86} \\
        VariantD (Ours) & \textsf{PC} & \textsf{CB} & \textsf{BS}  & 0.18 & 0.95 & \textbf{0.27} & 0.60 & \textbf{0.35} & \textbf{0.65} & 0.26 & 0.82 \\
        \bottomrule
    \end{tabular}
    \caption{ Ablation study of clustering (Cl) and pair sampling (Ps) in our concept extraction module. \textsf{NC}: Naive clustering, \textsf{PC}: Proposed clustering, \textsf{CB}: Concept Bank, 
    \textsf{BS}: proposed Bias-aware Sampling; \textsf{PS}: Proportional Sampling. 
    Data augmentation module in all variants is fixed. All results are obtained from Xception model trained on FF++.}     
    \label{tab:ablation_1}
\end{table*}

\begin{table*}[!ht]
    \centering
    \small
    \setlength{\tabcolsep}{4pt}
    \renewcommand{\arraystretch}{1.2}
    \begin{tabular}{lccccccccccc}
        \toprule
        \multicolumn{2}{c}{\multirow{2}{*}{Component}} & \multicolumn{8}{c}{Dataset} \\
        \cmidrule(l){3-10}
         &  & \multicolumn{2}{c}{FF++} & \multicolumn{2}{c}{DFDC} & \multicolumn{2}{c}{Celeb-DF} & \multicolumn{2}{c}{DFD} \\
        \cmidrule(lr){1-2} \cmidrule(lr){3-4} \cmidrule(lr){5-6} \cmidrule(lr){7-8} \cmidrule(l){9-10}
        Name & Da & $F_{\text{EO}}$ & AUC & $F_{\text{EO}}$ & AUC & $F_{\text{EO}}$ & AUC & $F_{\text{EO}}$ & AUC \\
        \midrule
        VariantA & \textsf{MU} & 0.27 & 0.94 & 0.32 & 0.56 & 0.33 & 0.63 & 0.27 & \textbf{0.80} \\
        VariantB &  \textsf{CM}  & \textbf{0.12} & \textbf{0.95} & 0.40 & 0.58 & \textbf{0.29} & 0.61 & 0.23 & 0.79 \\
        VariantC &  \textsf{FM}  & 0.27 & 0.89 & 0.41 & 0.57 & 0.44 & 0.59 & \textbf{0.22} & 0.78 \\
        VariantD (Ours) &  \textsf{PF}  & 0.15 & 0.94 & \textbf{0.29} & \textbf{0.60} & 0.30 & \textbf{0.68} & 0.27 & \textbf{0.80} \\
        \bottomrule
    \end{tabular}
    \caption{ Ablation study of data augmentation module (Da) in our framework. \textsf{CM}: CutMix, \textsf{MU}: MixUp, \textsf{FM}: Frequency Masking  \textsf{PF}: Proposed Frequency aware data augmentation. Concept extraction module in all variants is fixed. All results are obtained from the Xception model trained on FF++.  }
    \label{tab:ablation_2}
\end{table*}

\noindent\textbf{Effects of the proposed data augmentation.} We further investigate the performance improvement of our frequency-based CutMix method with that of other data augmentation methods.
The results in Table~\ref{tab:ablation_2} reveal the effects of our augmentation method are consistently better compared to other methods. 
Vanilla CutMix method (\textsf{CM}) severely degrades performance in AUC by 3\% on DFDC and 1\% on DFD.
We speculate that this is because the original \textsf{CM} method collapses deepfake-specific artifacts by combining images of different classes or mixing high-frequency components.
Similarly, MixUp based method \textsf{MU} also fails to enhance model fairness on FF++.
Frequency Masking \textsf{FM} often fails to improve the fairness metric except for DFD when comparing with our method.
This indicates that the diverse sampling in demographic attribute may be effective with respect to the model’s fairness generalization.
Spearman rank correlation tests comparing our method with others, yielded p-values less than 0.005 across all scenarios, thereby validating statistically significant improvements that the proposed method offers.
Overall, our data augmentation method yields the most substantial gains in fairness and AUC across all datasets.
%%%%%%%%%%%%%%%%%%%%%%%%%%%%%%%%%%%%%%%%%%%%%%%%%%%%%%%%%%%%%%%%%%%%%%%%%%%%%%%%

\section{Conclusion} 
We introduced a fairness-aware deepfake detection framework that employs temporal feature learning to identify demographic biases and frequency-aware data augmentation to mitigate them.
Through extensive experiments conducted on four large-scale deepfake datasets and two model architectures, we demonstrated the effectiveness of our approach in improving the fairness over existing methods while maintaining detection performance. 

A limitation of our method is its reliance on the assumption that deepfake-specific artifacts are predominantly present in the high-frequency domain. 
Thus, its effectiveness may be reduced in cases where forgery artifacts are distributed across the entire frequency spectrum.  
As part of future work, we will investigate the generalization capabilities of the proposed method when applied to SoTA classifiers beyond those considered in this study. 
We also aim to develop techniques that extend fairness-aware deepfake detection to speech and text-based forgery detection. And finally, we also plan to extend the method to detect deepfakes across non-face datasets.

{\small
\bibliographystyle{ieee}
\bibliography{egbib}
}

\end{document}